\documentclass[11pt,a4paper]{article}
\usepackage[hyperref]{emnlp2020}
\usepackage{times}
\usepackage{latexsym}

\usepackage{tikz}
\usepackage{multirow}
\usepackage{colortbl}

\usepackage{amsmath}
\usepackage{amssymb}
\usepackage{microtype}

\aclfinalcopy 

\title{Exploring Versatile Generative Language Model Via Parameter-Efficient Transfer Learning}
\author{Zhaojiang Lin\thanks{$^*$ Equal contributions.}, Andrea Madotto$^*$, Pascale Fung \\
Center for Artificial Intelligence Research (CAiRE)\\
  Department of Electronic and Computer Engineering\\
  The Hong Kong University of Science and Technology, Clear Water Bay, Hong Kong\\
\texttt{\{zlinao,amadotto\}@connect.ust.hk},\\ \texttt{ pascale@ece.ust.hk}}

\date{}

\begin{document}
\maketitle
\begin{abstract}
Fine-tuning pre-trained generative language models to down-stream language generation tasks has shown promising results.
However, this comes with the cost of having a single, large model for each task, which is not ideal in low-memory/power scenarios (e.g., mobile).
In this paper, we propose an effective way to fine-tune multiple down-stream generation tasks simultaneously using a single, large pre-trained model. 
The experiments on five diverse language generation tasks show that by just using an additional 2-3\% parameters for each task, our model can maintain or even improve the performance of fine-tuning the whole model~\footnote{Code available in \url{https://github.com/zlinao/VGLM}}.

\end{abstract}

\section{Introduction}
Large-scale language models~\cite{radford2019language,dai-etal-2019-transformer} have shown to be effective in learning highly transferable embedding, which can be used in several down-stream tasks. For instance, bidirectional models~\cite{peters2018deep, devlin2019bert} are fine-tuned to improve classification tasks~\cite{wang2018glue}, while, unidirectional language models~\cite{radford2019language} are more effective in language generation tasks. In this work, we focus on the latter, and show that it is possible to dynamically steer the output of a language model (e.g., GPT-2) towards a specific task (e.g., summarization) without modifying the original model parameters.


Feature-based transfer~\cite{howard2018universal} and fine-tuning~\cite{devlin2019bert} are the most commonly used methods for transfer learning of a language. The former freezes the pre-trained model and uses it as a feature extractor for training a new classifier, and the latter uses the pre-trained weight as a weight initialization for the model to be trained for downstream tasks. The feature-based transfer strategy has not shown promising results~\cite{devlin2019bert}, while fine-tuning, on the other hand, can achieve state of the art performance in multiple tasks~\cite{dong2019unified}. However, the downside of the latter is the need for a seperate model for each of the fine-tuned tasks. This is especially relevant for on-device applications, where a limited amount of computation/memory is available. 

Therefore, we study how to effectively use a single pre-trained model as the backbone for multiple language generation tasks, such as conversational question answering, summarization, machine translation, multi-turn chit-chat dialogue, and task-oriented natural language generation. This is a particular parameter-sharing schema, where we constrain the shared parameters to be the ones in the pre-trained model, and we learn task-specific parameters for each of the considered datasets. 

In this paper, we propose to use residual adapter layers~\cite{houlsby2019parameter} and task embeddings for modelling the aforementioned task-specific parameters, and we explore different training strategies such as distillation~\cite{hinton2015distilling,kim2016sequence}. We also analyse the trade-off between freezing or not freezing the language model parameters by leveraging two learning settings, multi-task (MT)~\cite{caruana1997multitask} and continual learning (CL)~\cite{thrun2012learning}. With our experiments, we empirically demonstrate that by adding less than 3\% task-specific parameters, our model can maintain or even achieve better performance than fine-tuning the whole model.

\section{Related work}
Pre-trained generative language models~\cite{radford2019language,radford2018improving,dai-etal-2019-transformer,yang2019xlnet,peters2018deep} have shown to be very effective in language generation, whereas, bidirectional pre-trained models~\cite{devlin2019bert,liu2019roberta,sanh2019distilbert} significantly improve the performance of several down-stream classification tasks. Fine-tuning large pre-trained models has shown positive results in dialogue tasks~\cite{DBLP:journals/corr/abs-1901-08149, budzianowski2019hello} and other language generation tasks~\cite{dong2019unified}. However, all of the previous works only consider fine-tuning on each generation task individually, which requires a separate model for each task. In this work, we use only a \textbf{single} model, for multiple generation tasks. 

Residual adapters, derived from residual networks~\cite{he2016deep}, were first introduced by \citet{rebuffi2017learning} for multiple visual domain learning.
\citet{houlsby2019parameter} proposed low-rank residual adapters to improve the scalability of the adapter module, and effectively transfer BERT~\cite{devlin2019bert} to multiple text classification tasks simultaneously, while \citet{bapna2019simple} applied an adapter layer to language/domain adaptation for neural machine translation. On the other hand,  \citet{dathathri2019plug} proposed a plug and play method to control the language model generation without finetuning the model. Differently, in this paper, we extend the idea of adapters to a large variety of language generation tasks, which has not been considered before, and we compare the idea of a fixed pre-trained back-bone for continual learning with multi-task training~\cite{stickland2019bert}.

\begin{table*}[t]
\centering
\resizebox{\textwidth}{!}{%
\begin{tabular}{r|c|cc|c|c|c|cc}
\hline
\multicolumn{1}{l|}{} & \multicolumn{1}{l|}{} & \multicolumn{2}{c|}{\cellcolor[HTML]{EE6666}\textbf{Persona (DLG)}} & \cellcolor[HTML]{F5A623}\textbf{NMT} & \cellcolor[HTML]{F8E71C}\textbf{SUM} & \cellcolor[HTML]{B8E986}\textbf{CoQA} & \multicolumn{2}{c}{\cellcolor[HTML]{50E3C2}\textbf{NLG}} \\ \cline{3-9} 
\multicolumn{1}{l|}{\multirow{-2}{*}{}} & \multicolumn{1}{l|}{\multirow{-2}{*}{\textit{Param.}}} & \textit{ppl.} $\downarrow$  & \textit{BLEU} $\uparrow$ & \textit{BLEU} $\uparrow$ & \textit{ROUGE 2} $\uparrow$ & \textit{F1} $\uparrow$ & \textit{BLEU} $\uparrow$ & \textit{AVG} $\uparrow$ \\ \hline

GPT-2 Finetune & 5$\times$ & \textit{\textbf{13.13}} & \textit{\textbf{2.17}} & \textit{\textbf{25.45}} & \textit{\textbf{18.1}} & \textit{\textbf{67.7}} & 66.4 & 57.77 \\ \hline
w/o Pre-Train & 5$\times$ & 37.77 & 0.99 & 16.52 & 17.0 & 15.1 & 60.5 & 53.51 \\ \hline
w/o Task Emb. & 5$\times$ & 13.24 & 0.00 & 0.61 & 15.0 & 35.2 & 53.1 & 47.25 \\ \hline
\hline
LM Head & 2.55$\times$ & 17.58 & 1.34 & 12.05 & 15.8 & 47.0 & 65.2 & 55.25 \\ \hline
VLM MT & 1.13$\times$ & \textbf{13.15} & 0.84 & 22.49 & 17.7 & \textbf{69.3} & 65.6 & 57.08 \\ \hline
VLM & 1.13$\times$ & 14.06 & \textbf{1.99} & \textbf{24.19}* & \textbf{18.0}* & 66.2 & \textbf{67.1} & \textbf{57.97} \\ \hline
w/o Task Emb. & 1.13$\times$ & 14.31 & 0.00 & 0.95 & 15.0 & 32.2 & 58.3 & 50.99 \\ \hline
\hline
Reference & - & 38.08$^{\mathparagraph}$ & - & 29.2$^\mathsection$ & 17.20 $^{\mathparagraph\mathparagraph}$ & 45.4$^{\dagger\dagger}$ & 65.9$^{\ddagger\ddagger}$ & 57.54 \\ \hline
SOTA & - & 17.51$^\dagger$ & - & 35.2$^\ddagger$  & 21.53 $^{\mathsection\mathsection}$ & 82.5$^\|$ & 66.2$^{\ddagger\ddagger}$ & 57.44 \\ \hline
\end{tabular} }

\caption{Results of VLM versus other fine-tuning techniques on the five evaluated datasets. Param. refers to the number of parameters that need to be stored after training. We use the adapter with distillation* for translation and summarization.
The Reference and SOTA results are: Profile Memory$^\mathparagraph$\cite{zhang2018personalizing}, TransferTransfo$^\dagger$~\cite{DBLP:journals/corr/abs-1901-08149}, DynamicConv$^\ddagger$\cite{wu2019pay}, Transformer$\mathsection$\cite{vaswani2017attention}, PG$^{\mathparagraph\mathparagraph}$ \cite{see2017get}, T5-11B$^{\mathsection\mathsection}$\cite{raffel2019exploring}, UniLM$^\|$\cite{dong2019unified}, PG$^{\dagger\dagger}$ \cite{reddy2019coqa} and SOTA system$^{\ddagger\ddagger}$ in \citet{dusek2019e2e} }\label{tab:res}
\end{table*}


\section{Methodology}
\begin{figure}[t]
    \centering
        \input{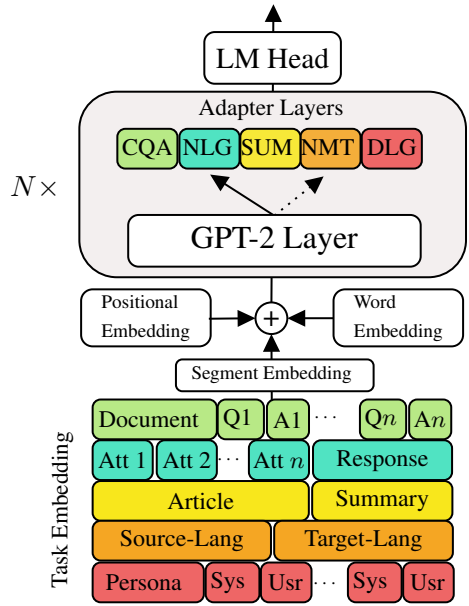}
    \caption{Simplified illustration of the Versatile Language Model. A detailed illustration is reported in Appendix A1.}
    \label{fig:my_label}
\end{figure}

The Versatile Language Model (\textbf{VLM}) is composed of three components: a pre-trained language model back-bone (e.g., GPT-2), and two kinds of specialized parameters for each generation tasks such as low-rank residual adapters and task embedding. Figure~\ref{fig:my_label} shows the VLM architecture with the specialized parameters in different colours.
\paragraph{Residual Adapters} These are trainable modules which steer the pre-trained model to different down-stream tasks. We adapt the design of the feed-forward Transformer sub-layer following~\citet{bapna2019simple}. To elaborate, the adapter block consists of 1) a layer normalization~\cite{ba2016layer} for an efficient adaptation and 2) a following autoencoder~\cite{hinton1994autoencoders}, with a residual connection. Formally, given the hidden representation $H_i \in \mathbb{R}^{t\times d}$ from the language model layer $i$, where $d$ is the hidden dimension and $t$ is the current generation step, the residual adapter computes the following:
\begin{equation}
\textrm{\textbf{Adapter}}(H_i) = (\textrm{\textbf{R}e\textbf{LU}} ( \textrm{\textbf{LN}}(H_i)W_i^{E} ))W_i^{D}  + H_i
\nonumber
\end{equation}
where $W_i^{E}$ and $W_i^{D}$ are parameters of dimension $d\times m$ and $m\times d$ respectively, and \textbf{LN}$(\cdot)$ denotes layer normalization. The bottleneck dimension $m$ is tunable and it allows to adjust the capacity of the adapter according to complexity of the target task. 




\paragraph{Task Embedding.} To adapt unconditional generative language models to different conditional language generation tasks (e.g., CoQA, Summarization), we construct a set of task-specific segment embeddings. For example, in multi-turn dialogue, we alternate between \textit{System} and \textit{User} embeddings to help the model to capture the hierarchical structure of dialogues. Figure~\ref{fig:my_label} shows the task embedding for each task, and more details are available in Appendix A2.

\paragraph{Knowledge Distillation} 
In tasks with a large distributional shift from the original pre-trained language model (e.g., Machine Translation), we expect a larger performance gap between VLM and full fine-tuning. To cope with this issue, we propose to use sentence-level knowledge distillation~\cite{kim2016sequence}, to help the task-specific parameters to better adapt to the task. Specifically, we first fully fine-tune a GPT-2 model on the training set of a task (e.g., Machine Translation). Then we replace the gold target (e.g., gold translation) in the training set with the greedy decoded output from the full fine-tuned model. Finally, the new constructed training set is used to fine-tune the student VLM.



\section{Experiments}

\subsection{Datasets \& Evaluation Metrics}
We conduct our experiment on five diverse datasets covering multiple generation tasks: Persona-Chat~\cite{zhang2018personalizing,dinan2019second} for chit-chat based dialogue (DLG), IWSLT~\cite{cettolo2016iwslt} German-English neural machine translation (NMT), CNN/Daily-Mail~\cite{hermann2015teaching,nallapati2016abstractive} for text-summarization (SUM), CoQA~\cite{reddy2019coqa} for generative conversational question answering (CQA), and E2E NLG-challenge~\cite{dusek2019e2e} for task-oriented natural language generation (NLG). 

We use a large variety of evaluation metrics, such as perplexity, F1 score, BLEU~\cite{papineni2002bleu}, ROUGE~\cite{lin2004rouge}, NIST~\cite{lin2004automatic}, METEOR~\cite{denkowski:lavie:meteor-wmt:2014} and CiDER~\cite{vedantam2015cider}. Each task uses the appropriate measure, as reported in Table 1, where in NLG we report the normalized average score of multiple metrics, as in \citet{dusek2019e2e}. More information about task description and the metrics used in each task are reported in Appendix A2. 

\subsection{Implementation and model comparison}
We implement VLM based on GPT-2-small (124M)~\cite{Wolf2019HuggingFacesTS}, and experiment with varying adapter bottleneck dimensions in $\{10, 50, 100, 300 \}$ and pick the best one in each task to trade-off the performance with the parameter efficiency. Specifically, we choose bottleneck sizes $100, 300, 100, 300$ and $10$ for DLG, NMT, SUM, QA, and NLG, respectively, which results in 13\% additional parameters in total. We ablate the adapter training with and without knowledge distillation and task embeddings. We also test the performance of a frozen back-bone (\textit{VLM}) to show the ability to continuously learn tasks, and multi-task fine-tune (\textit{VLM MT}) with a trainable backbone to show possible positive transferring among tasks as in ~\citet{stickland2019bert}. More training details and the dataset pre-processing are reported in Appendix A2.

To show the effectiveness of the proposed methodology of learning a versatile generative model, we compare (i) fine-tuneing the whole GPT-2 model for each task separately (\textit{GPT-2 Finetune}), (ii) fine-tuning the language model head of GPT-2 for each task (\textit{LM-Head}), (iii) existing baseline models reported (\textit{Reference}), and (iv) the state-of-the-art models for all the tasks (\textit{SOTA}).

\subsection{Results and Analysis}
Table~\ref{tab:res} shows the experimental results of the aforementioned models. Appendix A3 and A4 report detailed results and generated samples for all the datasets. Our findings can be summarized as follow: 

\paragraph{Fine-tuning GPT-2 vs Baseline \& SOTA.}
Fine-tuneing the whole GPT-2-small in each task can generally improve on the performance of competitive baselines such as Pointer-Generator~\cite{see2017get} in summarization (SUM) and CoQA. In both the Persona-Chat and the NLG tasks GPT-2 fine-tuning slightly outperforms the current SOTA, whereas, we observe a performance gap between GPT-2 and SOTA in NMT and SUM. Notably, the advantage of GPT-2 pre-training is limited in NMT: 1) no or little German text is present in the pretraining corpus; 2) the GPT-2 BPE~\cite{sennrich2016neural} tokenizer is optimized for English text, and not for multiple languages. Finally, in SUM and CoQA, the SOTA models use 100$\times$ bigger models~\cite{raffel2019exploring} and bidirectional attention~\cite{dong2019unified}, where instead, GPT-2 uses unidirectional attention. 


\paragraph{Adapter vs Fine-tuning GPT-2 \& LM Head.}
Fine-tuning only the adapter layers introduces 13\% additional parameters to the model with a minimal loss in performance (~$0.4$\%) compared to fine-tuning a seperate GPT-2 model. Moreover, the adapter layers are more effective, both in terms of performance and number of additional parameters, compared to fine-tuning LM-Head.

\begin{figure}[t]
    \centering
    \includegraphics[width=0.88\linewidth]{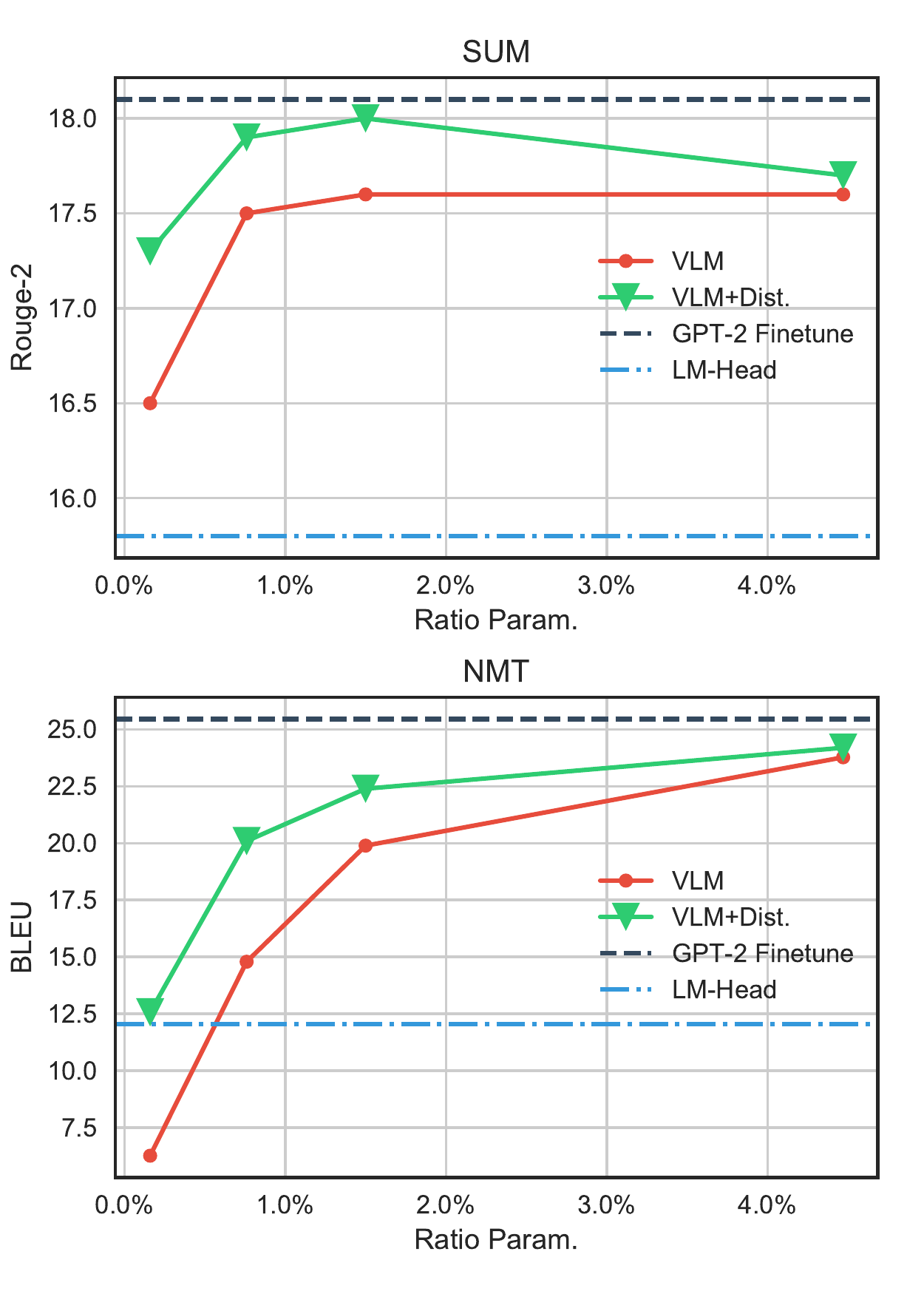}
    \caption{Performance comparison among different ratios of additional parameters for the SUM and NMT tasks.}
    \label{fig:nlg_nmt}
\end{figure}

\paragraph{Knowledge Distillation (KD)}
Using KD in the training procedure is especially useful in tasks such as NMT and SUM, where the gap between fine-tuning the whole model and adapter is large. This is because KD reduces the complexity of training targets (by replacing the gold training target with a teacher model generated target), which helps with low-capacity adapter (with 4\% parameter) by providing an easier translation/summarization task~\cite{zhou2019understanding}. Figure~\ref{fig:nlg_nmt} shows the effect of using distillation training when the gap with the full fine-tuning is more substantial. On the other hand, when the adapter performance is very close to that of the fine-tuning baseline, or better (i.e. NLG), distillation has no impact on the final performance. 

\paragraph{Task Embedding}
The specialized segment embedding (a.k.a. task embedding) is very important for achieving competitive performance, independently of the adapter. In Table~\ref{tab:res}, we can observe a substantial drop in performance when the task embedding is not deployed. Indeed, without a task embedding the model struggles to learn when the input sequence ends, and how to distinguish the different parts of the input sequence (e.g., attributes in NLG, document and question in CoQA etc.).

\paragraph{Frozen Backbone vs Trainable Backbone}
As previously mentioned, VLM can be trained either by freezing the weight of the GPT-2 model, i.e., independently and continually learning one task at a time, or by multitasking all the tasks and thus fine-tuning both the GPT-2 model and the adapters. The latter model has the advantage of being able to transfer knowledge among tasks, as we can observe in Table~\ref{tab:res} for the CoQA task, where VLM Multi-Task improve the F1 score by $3\%$. On the other hand, the frozen back-bone model has the big advantage of learning tasks sequentially, since the original GPT-2 weights remain untouched. 

\section{Conclusion}
In this paper, we have presented a Versatile Language Model which learns five diverse natural language generation tasks in a single model. We found that a residual adapter is more effective than fine-tuning other parts of the model (e.g., LM-Head), and that distillation helps in reducing the gap in performance in hard to fine-tune tasks, such as summarization and translation. Finally, we show the trade-off between a frozen and trainable back-bone, showing that the former has a competitive performance, with the advantage of being extendable to future tasks without full re-training. 

\bibliography{anthology,emnlp2020}
\bibliographystyle{acl_natbib}

\appendix
\clearpage
\section{Supplemental Material}
\label{sec:supplemental}

\subsection{Model details}
\begin{figure*}[t]
    \centering
    \resizebox{0.99\textwidth}{!}{
     \input{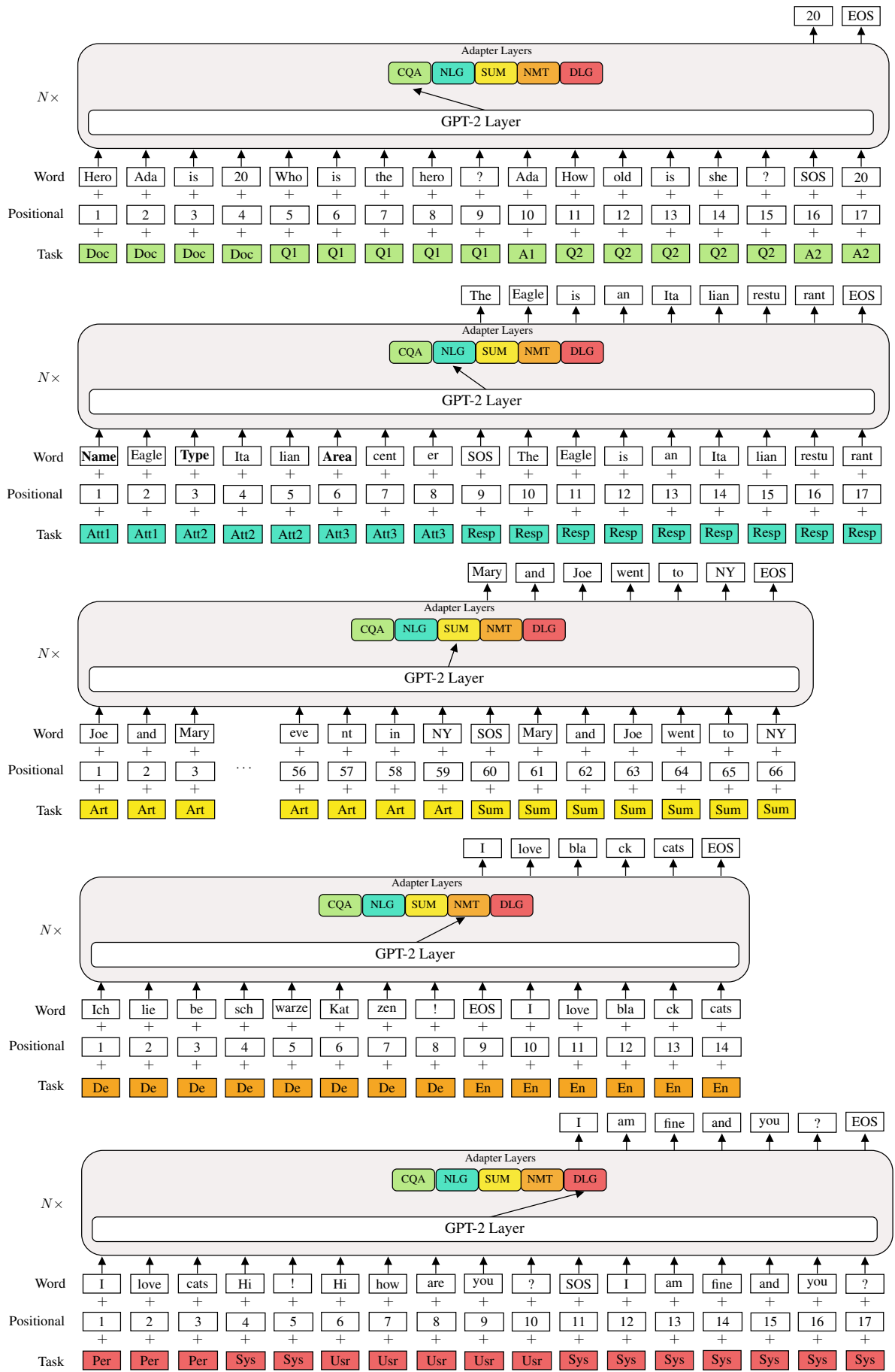}
     }
           
    \caption{A detailed version of VLM. VLM shares a GPT-2 back-bone and for each task, the model looks up a set of task embeddings and chooses the corresponding adapter.}
    \label{fig:Detail1}
\end{figure*}

Figure~\ref{fig:Detail1} illustrates a detailed version of VLM. VLM shares a GPT-2 back-bone and for each task, the model looks up a set of task embeddings for modeling different input structures and chooses the corresponding adapter.

\subsection{Experiment details}
In this section,  we will describe the dataset, evaluation metrics, dataset preprocessing and training details for each task.
\paragraph{Conversational Question Answering (CQA)}
CoQA~\cite{reddy2019coqa} is a free-form conversational question answering dataset. The task is to answer the questions in a conversation. Each turn in the conversation contains a question, and we need to answer the questions based on conversation histories and documents. We use \textit{document}, \textit{question}, and \textit{answer} segment embedding to help the model to distinguish the document and alternating questions and answers in the input sequence. We fine-tune the full GPT2-small or VLM (trainable adapter with a fixed GPT2-small) for five epochs with the Adam optimizer. For distillation we only fine-tune VLM for three epochs. We set the batch size to 16 and limit the maximum length of the document to 400 tokens and only retain the last two turns of questions and answers in the dialogue history. Following \citet{reddy2019coqa} we use the F1 score as evaluation metrics.

\paragraph{Summarization (SUM)}
CNN/Daily-Mail is a benchmark~\cite{hermann2015teaching,nallapati2016abstractive} for text summarization. We use \textit{article}, \textit{summary} segment embedding to divide the article and the summary. We fine-tune the full GPT2-small and VLM for 10 epochs with the Adam optimizer. For distillation, we only fine-tune VLM for five epochs. We set the batch size to 32 and limit the maximum length of the article to 400 tokens and that of the summary to 130 tokens. We use the ROUGE-1, ROUGE-2, and ROUGE-L scores~\cite{lin2004rouge} as evaluation metrics.

\paragraph{Neural Machine Translation (NMT)}
We use the spoken German-English translation dataset IWSLT~\cite{cettolo2016iwslt} as our NMT benchmark. We use \textit{source}, \textit{target} segment embedding to divide the source language and the target language. We fine-tune the full GPT2-small, VLM and distillated VLM for 8 epochs with the Adam optimizer. We set the batch size to 32 and limit the maximum length of the source and target sequence to 100 tokens. We use BLEU~\cite{papineni2002bleu} as the evaluation metric.

\paragraph{Persona Dialogue (DLG)}
The Persona-Chat dataset~\cite{zhang2018personalizing} is a persona-grounded multi-turn converstion dataset. We use \textit{persona}, \textit{system}, \textit{user} segment embedding to help the model to distinguish the persona, alternating system utterance and user utterance in an input sequence.  We fine-tune the full GPT2-small or VLM for three epochs with the Adam optimizer. We set the batch size to 16 and only retain the last five utterances in the dialogue history. We use perplexity, BLEU, and Consistency score~\cite{madotto2019personalizing} as evaluation metrics.

\paragraph{Natural Language Generation (NLG)}
The natural language generation challenge~\cite{dusek2019e2e} is a dataset for building a response generation module for task-oriented dialogue systems. Given a set of response attributes, the model needs to generate responses. For example, when the input attribute is \textit{name[The Wrestlers]}, \textit{priceRange[cheap]}, \textit{customerRating[low]}, the output should be \textit{The wrestlers offers competitive prices, but is not highly rated by customers}. We use a set of attribute segment embedding to segment the input attributes. We fine-tune the full GPT2-small and VLM for 10 epochs with the Adam optimizer. We set the batch size to 32 and use BLUE~\cite{papineni2002bleu} , ROUGE~\cite{lin2004rouge}, NIST~\cite{lin2004automatic}, METEOR~\cite{denkowski:lavie:meteor-wmt:2014} and CiDER~\cite{vedantam2015cider} as evaluation metrics.

\paragraph{Computational Cost}
Fine-tuning VLM requires around 80\%-90\% GPU memory compared to full-finetune the whole GPT-2 model, as it only updates the small ratio of parameters. And both models have similar training cost, we report the training speed with single GTX 1080 Ti:
\begin{table}[!ht]
\begin{tabular}{lcc}
\hline
\textbf{Task} & \multicolumn{1}{l}{\textbf{Training Speed}} & \multicolumn{1}{l}{\textbf{Training set size}} \\ \hline
\textbf{SUM}  & 7.5h/epoch                                  & 300, 000                                       \\
\textbf{NMT}  & 1.6h/epoch                                  & 200, 000                                       \\
\textbf{DLG}  & 1.5h/epoch                                  & 130, 000                                       \\
\textbf{QA}   & 5.0h/epoch                                  & 100, 000                                       \\
\textbf{NLG}  & 0.2h/epoch                                  & 42, 000                                        \\ \hline
\end{tabular}
\end{table}

\subsection{Detailed Results}
In this section, we report the detailed results for each task in Tables 2-6. We use a greedy decoding strategy for all the tasks.

\subsection{Example}

\begin{figure*}[t]
    \centering
    \includegraphics[width=\linewidth]{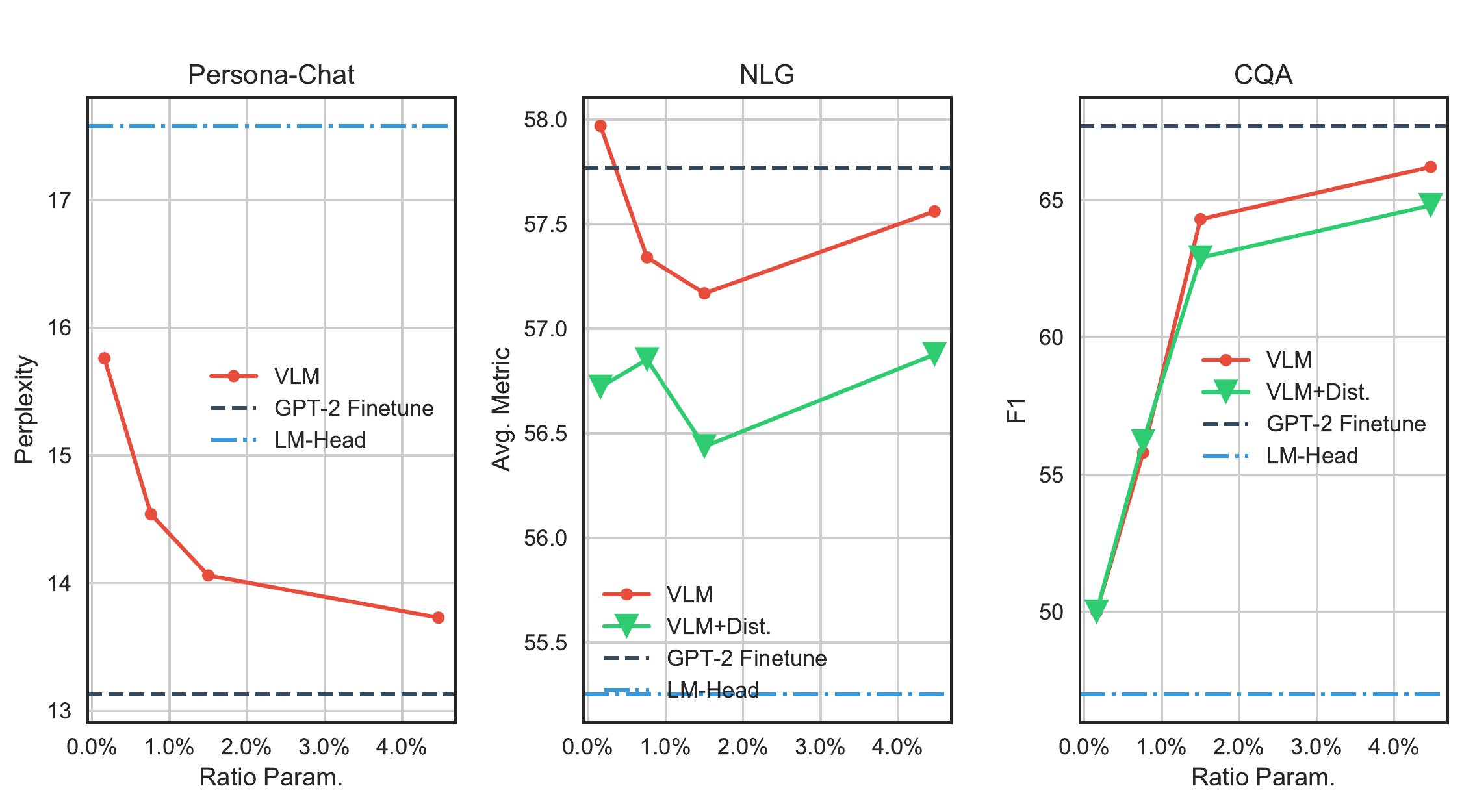}
    \caption{Performance comparison among different ratios of additional parameters. Here we can see that knowledge distillation does not improve the performance of the NLG task because of the small gap between VLM and the full fine-tuned GPT-2. Instead for the dialogue and QA tasks, the gold target is always better than the distillated target.}
    \label{fig:appendix}
\end{figure*}

\clearpage

\begin{table*}[t]
\begin{tabular}{|r|c|c|c|}
\hline
\rowcolor[HTML]{F8E71C} 
\multicolumn{4}{|c|}{\cellcolor[HTML]{F8E71C}\textbf{CNN / Daily Mail}} \\ \hline
\rowcolor[HTML]{F8E71C} 
\multicolumn{1}{|l|}{\cellcolor[HTML]{F8E71C}\textbf{Models}} & \textit{\textbf{ROUGE 1}} & \textit{\textbf{ROUGE 2}} & \textit{\textbf{ROUGE L}} \\ \hline
GPT Finetune & 37.4 & 18.1 & 27.7 \\ \hline
w/o Pre-Train & 35.5 & 17 & 26.2 \\ \hline
VLM mutli-task & 36.6 & 17.7 & 27 \\ \hline
VLM-10 (+ DIst.) & 35.0 (36.2) & 16.5 (17.3) & 25.0 (25.7) \\ \hline
VLM-50 (+ DIst.) & 36.4 (36.8) & 17.5 (17.9) & 26.6 (26.8) \\ \hline
VLM-100 (+ DIst.) & 36.5 (37.0) & 17.6 (18.0) & 27.0 (27.0) \\ \hline
VLM-300 (+ DIst.) & 36.6 (36.7) & 17.6 (17.7) & 26.6 (26.7) \\ \hline
PGNet~\cite{see2017get} & 39.53 & 17.28 & 36.38 \\ \hline
Bottom-Up~\cite{gehrmann2018bottom} & 41.22 & 18.68 & 38.34 \\ \hline
UniLM~\cite{dong2019unified} & 43.33 & 20.21 & 40.51 \\ \hline
T5-11B~\cite{raffel2019exploring} & 43.52 & 21.55 & 40.69 \\ \hline
\end{tabular}
\caption{Summarization results.}
\label{table:result1}
\end{table*}

\begin{table*}[t]
\begin{tabular}{|r|c|c|c|}
\hline
\rowcolor[HTML]{B8E986} 
\multicolumn{4}{|c|}{\cellcolor[HTML]{B8E986}\textbf{Persona}} \\ \hline
\rowcolor[HTML]{B8E986} 
\multicolumn{1}{|c|}{\cellcolor[HTML]{B8E986}\textbf{Models}} & \textit{\textbf{Perplexity}} & \textit{\textbf{BLEU}} & \textit{\textbf{Consistency (C)}} \\ \hline
GPT Finetune & 13.13 & 2.17 & 0.71 \\ \hline
w/o Pre-Train & 37.77 & 0.99 & 0.12 \\ \hline
VLM mutli-task & 13.15 & 0.84 & 0.27 \\ \hline
VLM-10 & 15.76 & 1.63 & 0.86 \\ \hline
VLM-50 & 14.54 & 1.84 & 0.72 \\ \hline
VLM-100 (+ DIst.) & 14.06 (89.34) & 1.99 (2.15) & 0.76 (0.72) \\ \hline
VLM-300 & 13.73 & 1.98 & 0.74 \\ \hline \hline
Deep Copy~\cite{yavuz2019deepcopy} & 54.58 & 4.09 & - \\ \hline
PAML-TRS~\cite{madotto2019personalizing} & 30.42 & 1.0 & 0.07 \\ \hline
ADAPT Centre (ConvAI2)~\cite{dinan2019second} & 29.85 & - & - \\ \hline
Persona-Chat~\cite{zhang2018personalizing} & 35.07 & - & - \\ \hline
TransferTransfero~\cite{wolf2019transfertransfo} & 17.51 & - & - \\ \hline
\end{tabular}
\caption{Persona Chat results.}
\label{table:result2}
\end{table*}

\begin{table*}[t]
\begin{tabular}{|r|c|}
\hline
\rowcolor[HTML]{EE6666} 
\multicolumn{2}{|c|}{\cellcolor[HTML]{EE6666}\textbf{CoQA}} \\ \hline
\rowcolor[HTML]{EE6666} 
\multicolumn{1}{|c|}{\cellcolor[HTML]{EE6666}\textbf{Models}} & \multicolumn{1}{l|}{\cellcolor[HTML]{EE6666}\textit{\textbf{F1}}} \\ \hline

GPT Finetune & 67.7 \\ \hline
w/o Pre-Train & 15.1 \\ \hline
VLM mutli-task & 69.3 \\ \hline
VLM-50 (+ DIst.) & 55.8 (56.2) \\ \hline
VLM-100 (+ DIst.) & 64.3 (62.9) \\ \hline
VLM-300 (+ DIst.) & 66.2 (64.8) \\ \hline
Seq2Seq~\cite{reddy2019coqa} & 27.5 \\ \hline
PGNet~\cite{reddy2019coqa} & 45.4 \\ \hline
DrQA~\cite{reddy2019coqa} & 54.7 \\ \hline
UNILM~\cite{dong2019unified} & 82.5 \\ \hline
Human~\cite{reddy2019coqa} & 89.8 \\ \hline
\end{tabular}
\caption{CoQA results.}
\label{table:result3}
\end{table*}

\begin{table*}[t]
\begin{tabular}{|r|c|}
\hline
\rowcolor[HTML]{F5A623} 
\multicolumn{2}{|c|}{\cellcolor[HTML]{F5A623}\textbf{NMT}} \\ \hline
\rowcolor[HTML]{F5A623} 
\multicolumn{1}{|c|}{\cellcolor[HTML]{F5A623}\textbf{Models}} & \textbf{BLUE} \\ \hline

GPT Finetune & 25.45 \\ \hline
w/o Pre-Train & 16.52 \\ \hline
VLM mutli-task & 22.49 \\ \hline
VLM-10 (+ DIst.) & 6.27(12.57) \\ \hline
VLM-50 (+ DIst.) & 14.79(20.09) \\ \hline
VLM-100 (+ DIst.) & 19.89(22.39) \\ \hline
VLM-300 (+ DIst.) & 23.77(24.19) \\ \hline
Transformer \cite{vaswani2017attention} & 29.2 \\ \hline
DynamicConv \cite{wu2019pay} & 35 \\ \hline
MIXER~\cite{ranzato2015sequence} & 21.83 \\ \hline
AC+LL~\cite{bahdanau2016actor} & 28.53 \\ \hline
NPMT~\cite{huang2017towards} & 28.96 \\ \hline
Dual Transfer Learning~\cite{wang2018dual} & 32.35 \\ \hline
LYC Transforemer~\cite{he2018layer} & 35.07 \\ \hline
\end{tabular}
\caption{NMT results.}
\label{table:result4}
\end{table*}

\begin{table*}[t]
\resizebox{\textwidth}{!}{%
\begin{tabular}{|r|c|c|c|c|c|c|}
\hline
\rowcolor[HTML]{50E3C2} 
\multicolumn{7}{|c|}{\cellcolor[HTML]{50E3C2}\textbf{NLG}} \\ \hline
\rowcolor[HTML]{50E3C2} 
\multicolumn{1}{|c|}{\cellcolor[HTML]{50E3C2}\textit{\textbf{Models}}} & \multicolumn{1}{l|}{\cellcolor[HTML]{50E3C2}\textit{\textbf{BLEU}}} & \multicolumn{1}{l|}{\cellcolor[HTML]{50E3C2}\textit{\textbf{NIST/10}}} & \multicolumn{1}{l|}{\cellcolor[HTML]{50E3C2}\textit{\textbf{METEOR}}} & \multicolumn{1}{l|}{\cellcolor[HTML]{50E3C2}\textit{\textbf{ROUGE L}}} & \multicolumn{1}{l|}{\cellcolor[HTML]{50E3C2}\textit{\textbf{CIDEr/10}}} & \multicolumn{1}{l|}{\cellcolor[HTML]{50E3C2}\textit{\textbf{norm. avg.}}} \\ \hline

GPT Finetune & 66.44 & 0.85279 & 0.4548 & 0.6911 & 0.22546 & 57.771 \\ \hline
w/o Pre-Train & 60.54 & 0.81697 & 0.4152 & 0.6471 & 0.19086 & 53.5106 \\ \hline
VLM mutli-task & 65.63 & 0.8342 & 0.4525 & 0.6889 & 0.22213 & 57.0806 \\ \hline
VLM-10 & 67.1 & 0.85046 & 0.4545 & 0.6935 & 0.229 & 57.9692 \\ \hline
VLM-50 & 66.01 & 0.84124 & 0.4568 & 0.6876 & 0.22128 & 57.3404 \\ \hline
VLM-100 & 65.38 & 0.83922 & 0.4564 & 0.6893 & 0.21972 & 57.1688 \\ \hline
VLM-300 & 66.18 & 0.84876 & 0.4539 & 0.6897 & 0.22387 & 57.5606 \\ \hline
VLM-10 + DIst. & 65.03 & 0.83199 & 0.456 & 0.6849 & 0.21286 & 56.721 \\ \hline
VLM-50 + DIst. & 65.23 & 0.83326 & 0.4576 & 0.6866 & 0.21287 & 56.8526 \\ \hline
VLM-100 + DIst. & 64.35 & 0.82485 & 0.4584 & 0.6852 & 0.20989 & 56.4368 \\ \hline
VLM-300 + DIst. & 65.19 & 0.83481 & 0.4575 & 0.6878 & 0.21182 & 56.8766 \\ \hline
TGEN baseline~\cite{dusek2019e2e} & 65.93 & 0.86094 & 0.4483 & 0.685 & 0.22338 & 57.5384 \\ \hline
SLUG~\cite{dusek2019e2e} & 66.19 & 0.8613 & 0.4454 & 0.6772 & 0.22615 & 57.44 \\ \hline
\end{tabular} }
\caption{NLG results.}
\label{table:result5}
\end{table*}

\begin{table*}[t]
\begin{tabular}{|r|l|}
\hline
\multicolumn{2}{|c|}{\textbf{NMT IWSLT 2014}} \\ \hline
\textit{Source} & \begin{tabular}[c]{@{}l@{}}Wenn ihr mit jemanden in den 20ern arbeitet, \\ einen liebt, wegen einem in den 20ern Schlaf \\ verliert, ich möchte euch seh en... O.k. Großartig.\\  Leute in den 20ern sind wirklich wichtig.\end{tabular} \\ \hline
\textit{GPT-2 Finetune} & \begin{tabular}[c]{@{}l@{}}If you work with somebody in the '20s, you love \\ them because you lost a loved one in the '20s, \\ I want to see you -- -- great. People in the '20s \\ are really important.\end{tabular} \\ \hline
\textit{VLM} & \begin{tabular}[c]{@{}l@{}}If you work with somebody in the '20s, because \\ of a love lost in the '20s, I want to see you -- OK. \\ Great. People in the '20s are really important.\end{tabular} \\ \hline
\textit{LM-Head} & \begin{tabular}[c]{@{}l@{}}If you work with someone in the 20ern, you love, \\ you love, you love, you love, you love, you love, \\ you love, you love, you love, you love, you love, \\ you love, you love, you love, you love, you love,\\ you love, you love, you love, you love, you love, \\ you love, you love, you love, you love, you love, \\ you love, you love, you love, you love, you love, \\ you love, you love, you love, you love, you love, \\ you love,\end{tabular} \\ \hline
\textit{Target} & \begin{tabular}[c]{@{}l@{}}If you work with twentysomethings, you love a \\ twentysomething, you're losing sleep over \\ twentysomethings, I want to see — Okay. Awesome, \\ twentysomethings really matter.\end{tabular} \\ \hline\hline
\textit{Source} & \begin{tabular}[c]{@{}l@{}}Ja, die Leute lassen sich später häuslich nieder als \\ früher, aber das machte Alex' 20er nicht zum \\ Entwicklungsausfall.\end{tabular} \\ \hline
\textit{GPT-2 Finetune} & \begin{tabular}[c]{@{}l@{}}Yes, people will be more domestic in the future \\ than they used to be, but that didn't make Alex' \\ 20s for failure.\end{tabular} \\ \hline
\textit{VLM} & \begin{tabular}[c]{@{}l@{}}Yes, people would come up later than they used \\ to, but that didn't make Alex' 20s a disaster.\end{tabular} \\ \hline
\textit{LM-Head} & \begin{tabular}[c]{@{}l@{}}Yes, people are later going to come back as former \\ former, but that doesn't make Alex' 20s anymore.\end{tabular} \\ \hline
\textit{Target} & \begin{tabular}[c]{@{}l@{}}Yes, people settle down later than they used to, \\ but that didn't make Alex's 20s a developmental \\ downtime.\end{tabular} \\ \hline \hline
\textit{Source} & Leute in den 20ern wie Alex und ich hatten nichts als Zeit. \\ \hline
\textit{GPT-2 Finetune} & People in the '20s like Alex and I didn't have time for time. \\ \hline
\textit{VLM} & People in the '20s like Alex and I had nothing but time. \\ \hline
\textit{LM-Head} & People like Alex and I had nothing as a time. \\ \hline
\textit{Target} & Twentysomethings like Alex and I had nothing but time. \\ \hline
\end{tabular}
\caption{NMT examples.}
\end{table*}

\begin{table*}[t]
\resizebox{\textwidth}{!}{%
\begin{tabular}{|r|l|}
\hline
\multicolumn{2}{|c|}{\textbf{Summurization CNN Daily Mail}} \\ \hline
\textit{Source} & \begin{tabular}[c]{@{}l@{}}los angeles -lrb- cnn -rrb- it's more than just one state's internal problem. \\ the historic california drought hurts the,rest of the union, too. that's \\ because california is a breadbasket to the nation, growing more \\ than a third of its vegetables and nearly two-thirds of its fruits and nuts. \\ here's why we should heed the ongoing drought in the most populous\\ state, a slowly expanding natural disaster now in its fourth year that \\ this week prompted gov. jerry brown to announce a,mandatory 25 \% ....\end{tabular} \\ \hline
\textit{GPT-2 Finetune} & \begin{tabular}[c]{@{}l@{}}the drought hurts the rest of the union, too, because california is a \\ breadbasket to the nation, the drought is, causing the nation's price \\ to rise, thanks to a strong dollar.\end{tabular} \\ \hline
\textit{VLM} & \begin{tabular}[c]{@{}l@{}}california drought hurts rest of the union, too, says jerry brown.\\ brown: prices rose last year for these items on your kitchen table. he \\ says prices expected to rise this year because of inflation.\end{tabular} \\ \hline
\textit{LM-Head} & \begin{tabular}[c]{@{}l@{}}new : the drought has hurt the rest of the union, the drought has hurt \\ the rest of the union, too, the drought has hurt the rest of the union, \\ too, the drought has hurt the rest of the union, too.\end{tabular} \\ \hline
\textit{Target} & \begin{tabular}[c]{@{}l@{}}americans paid more for some fruits and vegetables last year because \\ of the drought, tourists will now have to ask for a glass of water at a \\ california restaurant, perhaps the only good thing is another `` great '' \\ wine grape harvest last year.\end{tabular} \\ \hline \hline
\textit{Source} & \begin{tabular}[c]{@{}l@{}}-lrb- cnn -rrb- governments around the world are using the threat of terrorism \\ -- real or perceived -- to advance executions, amnesty international alleges \\ in its annual report on the death penalty. `` the dark trend of governments using \\ the death penalty in a futile attempt to tackle real or imaginary threats to state \\ security and public safety was stark last year, '' said salil shetty, amnesty's \\ secretary general in a release. `` it is shameful that so many states around the \\ world are essentially playing with people's lives -- putting people to death for \\ ` terrorism'or to quell internal instability on the ill-conceived premise of deterrence.\end{tabular} \\ \hline
\textit{GPT-2 Finetune} & \begin{tabular}[c]{@{}l@{}}report: governments are using the threat of terrorism to advance executions,\\ china is mentioned as having used, the death penalty as a tool in its `` strike hard ''\\ campaign., the number of executions worldwide has gone down by almost \\ 22 \% on the previous year.\end{tabular} \\ \hline
\textit{VLM} & \begin{tabular}[c]{@{}l@{}}amnesty international releases its annual report on the death penalty, the report \\ cites the example of pakistan, lifting a six-year moratorium on the execution \\ of civilians.\end{tabular} \\ \hline
\textit{LM-Head} & \begin{tabular}[c]{@{}l@{}}amnesty international : death penalty is a tool in state-sanctioned killing \\ campaigns, china is also mentioned, as having used the death penalty as a \\ tool in its `` strike hard '' campaign against terrorism.,china is also mentioned, \\ as having used the death penalty as a tool in its `` strike hard '' campaign \\ against terrorism\end{tabular} \\ \hline
\textit{Target} & \begin{tabular}[c]{@{}l@{}}amnesty's annual death penalty report catalogs encouraging signs, but setbacks \\ in numbers of those sentenced to death.,organization claims that governments \\ around the world are using the threat of terrorism to advance executions.,\\ the number of executions worldwide has gone down by almost 22 \% compared \\ with 2013, but death sentences up by 28 \%.\end{tabular} \\ \hline
\end{tabular}
}
\caption{SUM examples.}
\end{table*}

\begin{table*}[t]
\begin{tabular}{|r|l|}
\hline
\multicolumn{2}{|c|}{\textbf{Conversational QA (CoQA)}} \\ \hline
\textit{Source} & \begin{tabular}[c]{@{}l@{}}(CNN) -- Dennis Farina, the dapper, mustachioed cop turned-actor best \\ known for his tough-as-nails work in such TV series as "Law Order,"\\ "Crime Story," and  "Miami Vice," has died. He was 69.\\ "We are deeply saddened by the loss,of a great actor and a wonderful man,"\\ said his publicist, Lori De Waal, in a statement Monday. "\\Dennis Farina was always warmhearted and  professional, with a great\\ sense of humor and passion  for his profession. He will be greatly missed\\ by his family, friends and colleagues." Farina, who had a long career\\ as a police officer in Chicago, got into acting through director Michael Mann,\\ who used him as a consultant and cast him in his 1981 movie,"Thief." \\That role led to others in such Mann-created shows as  "Miami Vice"\\ (in which Farina played a mobster) and "Crime Story" (in which he\\ starred as Lt. Mike Torello).  Farina also had roles, generally as\\ either cops or gangsters, in a number of movies, including "Midnight \\ Run" (1988), "Get Shorty" (1995), "The Mod Squad" (1999) and\\ "Snatch" (2000). In 2004, he joined the cast of the long-running\\ "Law Order" after Jerry Orbach's departure, playing Detective \\Joe Fontana, a role he reprised on the spinoff "Trial by Jury." \\Fontana was  known for flashy clothes and an expensive car, a distinct \\ counterpoint to Orbach's rumpled Lennie Briscoe. Farina was on "Law Order"\\ for two years, partnered with Jesse L. Martin's Ed Green.\\ Martin's character became a senior detective after Farina left the show.\\ Q1: Is someone in showbiz?\\ A1: yes\\ Q2: Whom?\end{tabular} \\ \hline
\textit{GPT-2 Finetune} & Dennis Farina \\ \hline
\textit{VLM} & Dennis Farina \\ \hline
\textit{LM-Head} & Michael Mann \\ \hline
\textit{Target} & Dennis Farina \\ \hline
\end{tabular}
\caption{CQA examples.}
\end{table*}

\begin{table*}[t]
\begin{tabular}{|r|l|}
\hline
\multicolumn{2}{|c|}{\textbf{Conversational QA (CoQA)}} \\ \hline
\textit{Source} & \begin{tabular}[c]{@{}l@{}}Document:\\
Once upon a time, in a barn near a farm house, there \\ lived a little white kitten named Cotton. Cotton lived high up in \\ a nice warm place above the barn where all of the farmer's horses \\ slept. But Cotton wasn't alone in her little,home above the barn, \\ oh no. She shared her hay bed with her mommy and 5 other sisters. \\ All of her sisters were cute and,fluffy, like Cotton. But she was the \\ only white one in the bunch. The rest of her sisters were all orange \\ with beautiful white tiger stripes like Cotton's mommy. Being \\ different made Cotton quite sad. She often wished she looked like \\ the rest of her family. So one day, when Cotton found a can of the \\ old farmer's orange paint, she used it to paint herself like them. \\ When her mommy and sisters found her they started laughing. \\ "What are you doing, Cotton?!" "I only wanted to be,more like you". \\ Cotton's mommy rubbed her face on Cotton's and said "Oh Cotton, \\ but your fur is so pretty and special, like you. We would never want \\ you to be any other way". And with that, Cotton's mommy picked \\ her up and dropped her into,a big bucket of water. When Cotton \\ came out she was herself again. Her sisters licked her face until \\ Cotton's fur was all all dry. "Don't ever do that again, Cotton!" \\ they all cried. "Next time you might mess up that pretty white fur \\ of yours and we wouldn't want that!" Then Cotton thought, \\ "I change my mind. I like being special".\\
Q1: Whose paint was it?\\
A1: the farmer\\
Q2: What did Cotton's mother and siblings do when they saw \\ her painted orange?\end{tabular} \\ \hline
\textit{GPT-2 Finetune} & laughed \\ \hline
\textit{VLM} & laughed \\ \hline
\textit{LM-Head} & They put her in a big bucket of water \\ \hline
\textit{Target} & laughed \\ \hline
\end{tabular}
\caption{CQA examples.}
\end{table*}

\begin{table*}[t]
\begin{tabular}{|r|l|}
\hline
\multicolumn{2}{|c|}{\textbf{Dialogue Persona-Chat}} \\ \hline
\textit{Source} & \begin{tabular}[c]{@{}l@{}}\textbf{Persona}:\\i read twenty books a year\\               i'm a stunt double as my second job.\\               i only eat kosher. \\               i was raised in, a single parent household.\\
\textbf{Contex}t:\\USR: hello what are doing today?\\ SYS: i am good, i just got off work and tired, i have two jobs.\\ USR: i just got done watching a horror movie\end{tabular} \\ \hline
\textit{GPT-2 Finetune} & i love horror movies, i'm a stunt double. \\ \hline
\textit{VLM} & i love horror movies, i read twenty books a year. \\ \hline
\textit{LM-Head} & i love horror movies, i love them. \\ \hline
\textit{Target} & i rather read, i've read about 20 books this year. \\ \hline
\textit{Source} & \begin{tabular}[c]{@{}l@{}}\textbf{Persona}:\\ i'm a little girl.\\               i've superpowers.\\               i like to make time stop.\\               i live in the cloud.\\ \textbf{Context}:\\USR: hi i dye my hair 4 times a year\end{tabular} \\ \hline
\textit{GPT-2 Finetune} & that is cool. i like to dye my hair. \\ \hline
\textit{VLM} & wow that's a lot of dyeing. i love to dye. \\ \hline
\textit{LM-Head} & i love to make time stop \\ \hline
\textit{Target} & do you live in the sky? i do. in the clouds. \\ \hline
\end{tabular}
\caption{DLG examples.}
\end{table*}

\begin{table*}[t]
\resizebox{\textwidth}{!}{%
\begin{tabular}{|r|l|}
\hline
\multicolumn{2}{|c|}{\textbf{Natural Language Generation NLG}} \\ \hline
\textit{Source} & \begin{tabular}[c]{@{}l@{}}name = Blue Spice\\ eatType = coffee shop\\ near = Crowne Plaza Hotel\\ customer rating = 5 out of 5\end{tabular} \\ \hline
\textit{GPT-2 Finetune} & \begin{tabular}[c]{@{}l@{}}Blue Spice is a coffee shop near Crowne Plaza Hotel with a \\ 5 out of 5 customer rating.\end{tabular} \\ \hline
\textit{VLM} & \begin{tabular}[c]{@{}l@{}}Blue Spice is a coffee shop near Crowne Plaza Hotel. It has \\ a customer rating of 5 out of 5.\end{tabular} \\ \hline
\textit{LM-Head} & \begin{tabular}[c]{@{}l@{}}Blue Spice is a coffee shop near Crowne Plaza Hotel with \\ a customer rating of 5 out of 5.\end{tabular} \\ \hline
\textit{Target} & \begin{tabular}[c]{@{}l@{}}- The coffee shop Blue Spice is based near Crowne Plaza Hotel and \\ has a high customer rating of 5 out of 5.\\ - The Blue Spice coffee shop, near Crowne Plaza Hotel, has a \\ customer rating of 5 out of 5.\\ - If you want a coffee shop rated 5 out of 5 pick Blue Spice. It is \\ located near Crowne Plaza Hotel.\end{tabular} \\ \hline \hline
\textit{Source} & \begin{tabular}[c]{@{}l@{}}name = Giraffe\\ eatType = pub\\ familyFriendly = yes\\ food = English\\ near = Rainbow Vegetarian Café'\\ area =  riverside\end{tabular} \\ \hline
\textit{GPT-2 Finetune} & \begin{tabular}[c]{@{}l@{}}Giraffe is a family friendly pub located near Rainbow Vegetarian Café. \\ It serves English food and is located in t\\ he riverside area\end{tabular} \\ \hline
\textit{VLM} & \begin{tabular}[c]{@{}l@{}}Giraffe is a family friendly pub located near Rainbow Vegetarian Café \\ in the riverside area.\end{tabular} \\ \hline
\textit{LM-Head} & \begin{tabular}[c]{@{}l@{}}Giraffe is a pub near Rainbow Vegetarian Café that serves English food. \\ It is family friendly and located in the,riverside area.\end{tabular} \\ \hline
\textit{Target} & \begin{tabular}[c]{@{}l@{}}- The Giraffe,,a child-friendly, English pub, is,located in riverside area, \\ near Rainbow Vegetarian Café.\\ - Giraffe is a pub that is children friendly and,provides English food.It is \\ located on the riverside near Rainbow Vegetarian Café.\\ - There is an English based pub called Giraffe. It is located in the riverside \\ area near the Rainbow Vegetarian Café and, yes, it is kid friendly.\end{tabular} \\ \hline
\end{tabular}
}
\caption{NLG examples.}
\end{table*}

\end{document}